\documentclass[10pt,fleqn,conference]{IEEEtran}

\usepackage{cite}
\usepackage{amsmath}
\usepackage{amsfonts}
\usepackage{algorithmic}
\usepackage[caption=false]{subfig}
\usepackage{graphicx}
\usepackage{float}

\begin{document}
\title{GAN Augmented Text Anomaly Detection with Sequences of Deep Statistics\\}
\author{\IEEEauthorblockN{Mariem Ben Fadhel}
\IEEEauthorblockA{School of Electrical and\\Computer Engineering\\
Morgan State University\\
Email: maben5@morgan.edu}
\and
\IEEEauthorblockN{Kofi Nyarko}
\IEEEauthorblockA{School of Electrical and\\Computer Engineering\\
Morgan State University\\
Email: kofi.nyarko@morgan.edu}}

\maketitle

\begin{abstract}Anomaly detection is the process of finding data points that deviate from a baseline. In a real-life setting, anomalies are usually unknown or extremely rare. Moreover, the detection must be accomplished in a timely manner or the risk of corrupting the system might grow exponentially. In this work, we propose a two level framework for detecting anomalies in sequences of discrete elements.  First, we assess whether we can obtain enough information from the statistics collected from the discriminator's layers to discriminate between out of distribution and in distribution samples. We then build an unsupervised anomaly detection module based on these statistics. As to augment the data and keep track of classes of known data, we lean toward a semi-supervised adversarial learning applied to discrete elements.
\end{abstract}

% no keywords

\section{Introduction}
% no \IEEEPARstart

The human immune system works in a highly adaptive mechanism. It's comprised of the first line of defense,  a generic innate system that is capable of taking action quickly to protect the body from outside pathogens. And the second line of defense, the adaptive immune system, which is slower but has a long time memory, and is much more specialized and more efficient than the innate defense system if the antigen is known.\\

We define an anomaly as a data point that deviates significantly from the normal data distribution. In this context, we consider a data point as a sequence of discrete tokens. The problem of anomaly detection is challenging because only benign data is known while outliers are rare to unknown. The ability of efficiently detecting novelties is of a great use in a plethora of domains: network security, image denoising. \cite{Buades05anon-local}, detecting alterations in texts\cite{kumar2015vews}, in finance for auditing\cite{DBLP:journals/corr/abs-1709-05254}, etc..
The anomaly detection process must start by modeling the representation of the baseline by efficiently identifying the variations that define the positive class.\\

In this work we aim to solve the problem of anomaly detection in sequences of discrete data, text data more specifically. In addition to the discreteness of data, text adds few more challenges; features are mostly unknown, multidimensional and we need to account for the context of each word to grasp the semantic meaning. The framework presented can be applied to solve problems like code authorship analysis, code injections and bot identification in social media for example.

Another challenge is how to take advantage of the scarce novel/malicious data samples so that it's recognized in the next run of the outlier detection system.\\

Generative adversarial networks (GANs) are methods that aim to generate realistic samples similar to the real data by learning statistical properties close enough to the real data distribution. It is done by training a generator and a discriminator in a zero-sum game setting and minimizing the Jenson-Shannon Divergence (JSD) between the distribution of the real data and the generated data. \cite{NIPS2014_5423}
\begin{equation}\label{eq:lossgantextgan}
\ \mathcal{L}_{GAN} = \mathbb{E}_{s\sim\mathcal{S}}logD(s) + \mathbb{E}_{z\sim{p_{z}}}log[1-D(G(z))]
\end{equation}

The standard version of GANs is hard to train and often suffers from convergence problems such as mode collapse and vanishing gradient [5], [6]. Semi-supervised GANs alleviate some of these shortcomings and present a more stable version of GANs. It's capable of generating better samples thanks to the introduction of unlabeled data that act as a regularizer. Semi-supervised GAN are capable of classifying data and also generating comparable samples to real data which could be of great use in the context of anomaly detection. It can be leveraged by recognizing known classes of data-- we set the known classes to be not only the baseline classes but also known or
previously encountered rare classes of malicious data-- and augmenting these classes by generating additional samples to optimize the classification boundaries. In fact \cite{NIPS2017_7229} demonstrates that the generator in a semi-supervised GAN generates data from a mixture distribution of all these classes. The challenge remaining is to alter their objective function and render them workable on discrete data in a semi-supervised context.
The first line of defense, more comparable to the innate human immune system, aims to recognize out-of-distribution data that we consider anomalous. It receives a sequence of deep statistics and decides solely with batches of baseline data if it's in fact anomalous. If the sample is from a known distribution to the system (in-distribution) the second line of defense takes over by providing the right class corresponding to the sample.\\

Our contributions are in three folds: (1) In the second section, We present a formulation of a conditional adversarial learning objective adequate for discrete sequences for data augmentation and classification. (2) The third section provides an analysis of the out-of-distribution detection module using information derived from variations in the deep layers' statistics. (3) We propose a framework for text anomaly detection in the fourth section. A description of the experiments is provided in the fifth section, followed by the main results. We mention the related work in the sixth section and present a conclusion in the last section.

\section{Conditional Semi-supervised Text Gan}
GANs were created initially to generate continuous samples and exhibit limitations when it comes to discrete data, as it hinders the back propagation process. The reason being that gradient updates from the discriminator will not necessarily match a value in the discrete domain during the backward propagation of the gradients from the discriminator to the generator which is only feasible if the generated samples comes from a continuous distribution. A growing body of literature has investigated this problem. \cite{Yu2016},\cite{NIPS2017_6908},\cite{DBLP:journals/corr/abs-1709-08624} proposed solutions based on reinforcement learning and the REINFORCE algorithm turning the problem into a sequential decision making process. The Monte Carlo search is used to estimate the discrete tokens.
A different approach aims to optimize the discrete sequences by providing continuous approximations through the adoption of the Gumbel-Softmax trick\cite{45822} or the concrete distribution\cite{DBLP:journals/corr/MaddisonMT16} like in\cite{DBLP:journals/corr/KusnerH16}.

TextGAN \cite{zhang2017adversarial} is also part of the non-reinforcement approach, it uses the soft-argmax to approximate . It's a framework that aims to generate realistic sentences with a generative adversarial network. A Convolutional Neural Network (CNN) discriminator is used to extract salient features from sentences through a process of convolving filters to word embeddings of different window sizes. Filters extract a different linguistic features by varying the window size. Convolution operations are followed by max-over-time pooling extracting the most important features from feature maps.A softmax layer allows to decide if the final feature vector belongs to the generator or the discriminator's distribution. A Long-Short Term memory neural network (LSTM) is used as a generator. All words in the generated sentence are created using the words generated until the end of sentence symbol is generated. The input of the $t^{th}$ step is the embedding of the generated word at step $ t-1 $.

The objective function is inspired by the feature matching objective in \cite{DBLP:journals/corr/SalimansGZCRC16} and is defined below:
\begin{equation}\label{eq:lossgtextgan}
\ \mathcal{L}_{G} = \mathcal{L}_{MMD^{2}} 
\end{equation}
\begin{equation}\label{eq:lossdtextgan}
\ \mathcal{L}_{D} = \mathcal{L}_{GAN} -\lambda_{r}\mathcal{L}_{recon}+\lambda_{m}\mathcal{L}_{MMD^{2}}
\end{equation}
\begin{equation}\label{eq:reconloss}
\ \mathcal{L}_{recon} = \|{\hat{z} - z}\|^2
\end{equation}\\

Instead of matching the real sentences, the generator attempts to match the synthetic sentence features to the real sentence features by minimizing the MMD between the two distributions \eqref{eq:lossdtextgan}.\\
$\mathcal{L}_{recon}$ is the reconstruction loss which is the distance between the latent variable and its reconstructed version.
The discriminator's loss \eqref{eq:lossdtextgan} incorporates the standard GAN loss \eqref{eq:lossgantextgan} which renders the model vulnerable to mode collapse. In a comparative study \cite{DBLP:journals/corr/abs-1802-01886} provides an analysis where it proves that TextGAN is, in fact, prone to mode collapse.

On the other hand, semi-supervised generative adversarial learning has proven to produce more stability in training and to provide a significant improvement in regards to the quality of generated samples. \cite{journals/corr/Springenberg15} argues that catGAN a categorical generative adversarial network with a multi-class classifier as a discriminator, has a regularization effect on its discriminator.\\
Given a corpus of sequences $\mathcal{S}$, a labeled set of sequences $\mathcal{LS}={(s,y)}$ and $\mathcal{Y} = {1,2,...,K}$ label space for classification with K being the number of classes, let $P_{D}$ be the distribution corresponding to the discriminator and $P_{G}$ be the distribution corresponding to the generator. \\
Inspired by the discriminator's loss function in a semi-supervised generative adversarial network \cite{DBLP:journals/corr/SalimansGZCRC16}, we define the discriminator loss of a GAN over sequences of discrete elements as: 
\begin{equation}\label{eq:lossdtextsslgan}
	\ \mathcal{L}_{D} = \mathcal{L}_{DSSL} -\lambda_{r}\mathcal{L}_{recon}+\lambda_{m}\mathcal{L}_{MMD^{2}}
\end{equation}
\begin{equation}\label{eq:ssldloss}
	\begin{split}
		\mathcal{L}_{DSSL} = {}& \mathbb{E}_{s,y\sim\mathcal{LS}}logP_{D}(y|s,y\leq{K}) + \\
		&\mathbb{E}_{s\sim\mathcal{S}} logP_{D}(y\leq{K}) + \\ &\mathbb{E}_{\tilde{s}\sim{G}}logP_{G}(K+1|\tilde{s})
	\end{split}
\end{equation}

The first term in \eqref{eq:ssldloss} is the log conditional probability for labeled sequences. The second term is the log probability for the K classes. Notice that it's not conditioned because it concerns the unlabeled sequences. And finally, the last term is the conditional log probability of generated data, $\tilde{s}$ being the synthetic sequence generated by G.\\
In order to optimize the objective function, the discriminator's loss function   in \eqref{eq:lossdtextsslgan}, is maximized and the generator's loss function in \eqref{eq:lossgtextgan} is minimized.

\section{The novelty detection module}
Neural networks are easily misled by adversarial examples due to their linearity. Those are data points that had been perturbed as to incur an erroneous classification with high confidence\cite{43405}. This fact is rather alarming in a world that is increasingly relying on artificial intelligence and machine learning for crucial tasks. Results obtained in \cite{DBLP:journals/corr/HendrycksG16c} show that by analyzing the statistics of the softmax output probabilities of inference samples, it's possible to distinguish out-of-distribution samples and to predict a misclassification.\\

We propose to analyze the patterns of sequences collected from the output statistics at each layer of the discriminator and aim to ascertain whether it provides a resilient detection of out-of-distribution samples.
Let $V_{l}$ be the vector output of layer l, $V_{l} = logit_{l}$ with values from logits at layer l. Let $SV _{s}$ be the sequence of vectors $V_{l}$ for a data sample s from $\mathcal{S}$. We train an LSTM Autoencoder neural network on batches of sequences $SV _{s}$. Autoencoders are feed forward neural networks that are trained to learn the most important features that lead to generating an almost identical copy of the input. Its objective is to minimize the loss of reconstructing the input using back-propagation, called the reconstruction error. A high reconstruction error of a sample signals that sequence of statistics is anomalous. The threshold marking the limit of acceptable reconstruction error can also be learned based exclusively on baseline samples. An advantage of Autoencoders is that the deviation from benign input is possible without the introduction of malicious data which fits the problem of anomaly detection. 
The Autoencoder framework had been extensively used for anomaly detection \cite{Zhou:2017:ADR:3097983.3098052},\cite{7823921},\cite{Yan2015OnAA},\cite{10.1007/978-3-319-48057-2_9}.\\

\section{Text Anomaly Detection Framework}
\begin{figure}[ht!]
    \centering
    \includegraphics[width=\textwidth, height=.10\textheight, keepaspectratio, trim=3cm 15cm 4cm 8cm, clip]{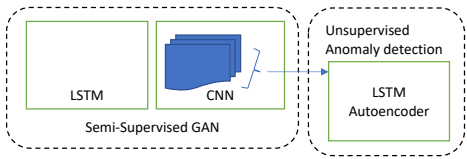}
    \caption{Framework for Text Anomaly Detection}
    \label{fig:AD_Framework}
\end{figure}

The text anomaly detection framework works as follows:
The semi-supervised GAN is trained on baseline data. Sequences of statistics of each layer of the discriminator are saved and prepared (normaized and scaled) to train the LSTM Autoencoder. 
The LSTM Autoencoder is trained and validated using a portion of the statistics' sequences. At this point, the reconstruction error threshold is adjusted.\\
In Inference mode, the process starts with the LSTM Autoencoder. To check if a data sample is malicious, the reconstruction error of the autoencoder is calculated and compared to the threshold. If it's above it, then the data sample is an out-of-distribution sample and there is no need to go further. It's tagged as anomalous. If it's less than the threshold, we inject the last column that corresponds to the softmax layer into the last layer of the discriminator to get the class it corresponds to. The figure below ( Fig.\ref{fig:AD_Framework}) depicts the components of the text anomaly detection framework. 

\section{Experiments}
The anomaly detection model is trained on two datasets, namely The sentence polarity dataset\cite{Pang+Lee:04a}, which consists of 1000 positive and 1000 negative movie reviews, and 20Newsgroups dataset comprised of 20,000 news group documents. It's a popular dataset for machine learning experiments on text data.  
We start by training a Convolutional Neural Network. We use the standard splitting of data into training, validation and testing sets with a ratio of 0.6, 0.2, 0.2 respectively. 20Newsgroups is set as the baseline whereas the polarity dataset is the malicious dataset. We only make use of the latter as to test the model, it's not part of the training process, and the anomaly detection method doesn't depend on it.\\

After training the Generative adversarial network on both datasets, two news datasets are built from the discriminator network's layers. We use the same architecture as \cite{zhang2017adversarial}. Text inputs are vectorized using GLOVE \cite{pennington2014glove} to obtain word embedding which generates a considerable size of the logits at the embedding layer. For that reason, and because of limited computational resources, we decided to drop the embedding layer's output from the sequences of statistics.
Scaling, normalization and splitting is performed on the sequences using it to train the LSTM Autoencoder. 

\section{Results}
\begin{figure}[!ht]
    \centering
        \subfloat[Baseline data\label{subfig-1:re}]{
        \includegraphics[width=0.45\columnwidth]{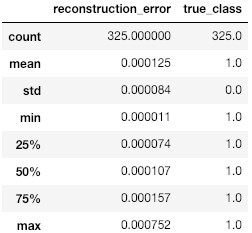}
        }
        \hfill
        \subfloat[Novel data\label{subfig-2:re2}]{
        \includegraphics[width=0.45\columnwidth]{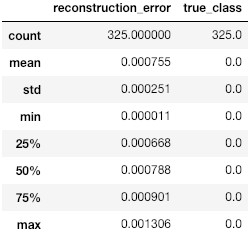}
        }
    \caption{LSTM Autoencoder reconstruction errors for normal and anomalous statistics.}
    \label{fig:reconst_ae}
\end{figure}

Evaluating autoencoders, is not as straight forward as evaluating other neural networks, like classifiers for example. Its performance depends on how well it the task it's used for, so the evaluation process should be designed accordingly. In this work, we are interested in quantifying the performance of the model in discriminating data samples that are anomalous; coming from a distribution other than the one it was trained on. Most of the standard evaluation metrics evaluate classifiers and require at least two classes of labels. Accuracy is the fraction of the predictions that were correct. In regression tasks, it doesn't give an accurate judgment and other measures should be investigated in tandem. The hyperbolic tangent activation function is preferred to relu for this task, and produces a higher accuracy of the model. More important for us is to look at the progression of the loss on training and evaluation batches to make sure that the model is learning properly but not over-fitting the data.
The reconstruction error is the metric that will give us insight on whether the model is fitting the data it was trained on. Out-of-distribution samples must trigger a higher reconstruction error if the sequences of statistics are informative of anomalous samples, and that will confirm our assumption.
We show in the two tables below the reconstruction error for in and out of distribution sequences of statistics. 
Reconstruction errors Fig. \ref{fig:reconst_ae} are much higher for sequences coming from out-of-distribution data than for baseline data.

\begin{figure}[ht!]
    \centering
    	\subfloat[Training loss and validation loss thought epochs.\label{subfig-1:loss}]{
        	\includegraphics[width=0.45\columnwidth]{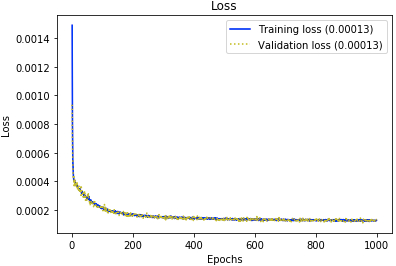}
    }
    \hfill
    	\subfloat[Recall vs. precision. \label{subfig-2:recall_precision}]{
        \includegraphics[width=0.45\columnwidth]{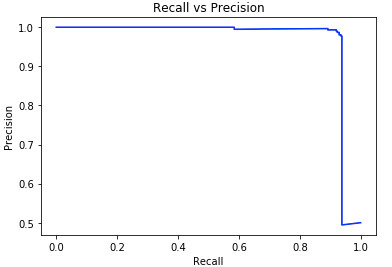}
    }
	\hfill
   	\subfloat[Area Under the Receiver Operating Characteristic curve. \label{subfig-3:auc}]{
        \includegraphics[width=0.45\columnwidth]{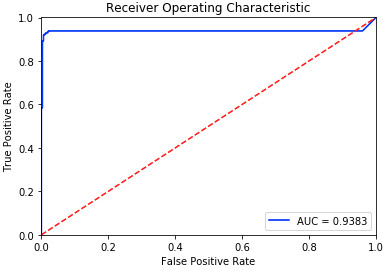}
    }
	\hfill
    \subfloat[Recall vs Reconstruction error. \label{subfig-4:recall_re}]{
    \includegraphics[width=0.45\columnwidth]{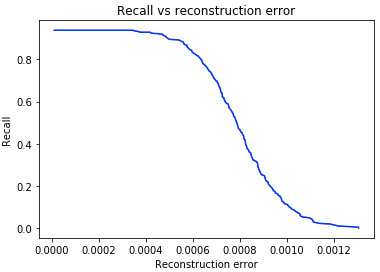}
    }
\caption{LSTM Autoencoder performance analysis for anomaly detection of sequences of the discriminator's training statistics.}
\label{fig:mesure_ae}
\end{figure}

The autoencoder is trained in an unsupervised way, but for the sake of validating our assumption, we labeled the sequences of the statistics collected from each layer in the discriminator's network and that we trained the LSTM autoencoder on as one class, the ones coming from a different distribution as another.
The receiver operating curve is used to visualize the performance of a classifier. It shows the true positive rate versus the false positive rate under different thresholds. A more steep curve goes along with a better performance. 
The recall and precision are the  defined as:\\
\begin{equation}\label{eq:precision}
\ Precision =  \dfrac{True Positives}{True Positives + False Positives}
\end{equation}
\begin{equation}\label{eq:recall}
\ Recall = \dfrac{True Positives}{True Positives + False Negatives}
\end{equation}
A high value of recall and precision is what we aim for, and it means that many correct predictions are returned and have high relevance.
We also plot the recall with respect to the reconstruction error under different thresholds. This plot gives us a fair idea on the optimal value of threshold to detect anomalous samples. Based on that threshold we visualize the confusion matrix to test if the threshold is adequate in generating an acceptable number of erroneous predictions. Fig. \ref{fig:mesure_ae} shows the results of the analysis previously discussed.

\section{Previous Work and Discussion}
The problem of applying GAN networks to anomaly detection is attracting an increasing interest. GANs are superior to other generative models like autoencoders or variational autoencoder in producing realistic data.\cite{NIPS2017_7137}. Augmenting the baseline data through the generator and training a classifier to recognize real from fake data is tempting to apply the model in the context of anomaly detection. The pitfall though is that the fake data class (K+1), K being the first labeled classes (K could be 1 in the case of a binary classifier), does not represent all the unknown distributions related to the anomalous data. Also, the generator tends to match the data distribution of real samples, and in the case of feature matching\cite{DBLP:journals/corr/SalimansGZCRC16}, the space of real data features. It means that the fake distribution tends to get closer as optimizing the GAN.\\

There are mainly three approaches adopted for this problem. First is leveraging the discriminator and the generator both to conduct anomaly detection.\cite{Schleglb} presents AnoGAN an anomaly detection scheme for anomalous image detection based on identifying disease markers. They provide two scoring schemes one based on a residual loss of the distance between a real image and a generated image and a feature matching discrimination loss that computes the loss of the discriminator output on when fed a generated image. This allows deciding if an image comes from the generator distribution by a process of inverse mapping. They conclude that the residual loss is enough for the anomaly detection task.\\

In\cite{Wang2018AnomalyDV} proposed a minimum likelihood method to force the generator to produce values that are distant from the normal distribution which is counter-intuitive to the goal of GAN which is to make the generator produce data similar to the real one. Another work that leverages GANs for fraud detection is \cite{Zheng2018OneClassAN}, they train a complementary generator based on the work from \cite{NIPS2017_7229}. In \cite{NIPS2017_7229} the complementary generator produces data samples in low-density areas of the data distribution and is used as a generalization method and a solution to optimize learning, this is not quite representative of anomalous data.\\

\section{Conclusion}
In this work we present a novel approach to GAN based anomaly detection, we define it in the context of semi-supervised learning to allow an adaptive framework that gets better at recognizing anomalous data with experience.
We present promising results when applying our out-of-distribution component to two different data sets of sequences of discrete data. We are interested in the future in analyzing the performance of our method when incorporating the Movers distance measure \cite{Chen2018AdversarialTG} in the definition of the discriminator loss.

\bibliographystyle{./IEEEtran}
\bibliography{./IEEEabrv,./IEEEexample,./GANSAD}

\end{document}